\documentclass[12pt]{report}

\usepackage{graphicx}

\usepackage{graphicx}

\usepackage{xspace}

\usepackage{amsopn}

\usepackage{amsfonts}

\usepackage{color}

\usepackage{graphicx}

\usepackage{epsfig}

\usepackage{epstopdf}

\usepackage{amsthm}

\usepackage{float}

\usepackage{multirow}

\usepackage{caption}

\usepackage{subcaption}

\usepackage{amssymb}

\usepackage{tabularx}

\usepackage{rotating}

\usepackage{amssymb,amsmath,bm}

\usepackage{epstopdf}

\usepackage{multirow}

\usepackage{amssymb,mathrsfs}

\usepackage{epstopdf}

\usepackage{amssymb,mathrsfs}

\usepackage[]{algorithm2e}

\DeclareMathAlphabet{\mathpzc}{OT1}{pzc}{m}{it}

\usepackage[lmargin=3cm,rmargin=3cm,tmargin=3cm,bmargin=3.5cm]{geometry}

\usepackage{nomencl}

\author{Name}

\date{}

\DeclareMathAlphabet{\mathpzc}{OT1}{pzc}{m}{it}
\begin{document}


\thispagestyle{empty}

\begin{large}

\begin{center}

{\bf }

{\Large{\textbf{\sc SOCIAL MEDIA ANALYSIS BASED ON SEMANTICITY OF STREAMING AND BATCH DATA}\\

\vspace{.3cm}

}}

\vspace{1cm}

A THESIS \\

\vspace{.3cm}

\textit{Submitted by}\\

\vspace{.3cm}

\textbf{BARATHI GANESH HB}\\

\textbf{(CB.EN.P2CEN13019)}\\

\vspace{.3cm}

\textit{in partial fulfillment for the award of the degree of}\\

\vspace{.5cm}

\textbf{MASTER OF TECHNOLOGY}\\

\textbf{IN}\\

\textbf{COMPUTATIONAL ENGINEERING AND NETWORKING}\\

\begin{figure}[h]

\hspace*{5cm}\includegraphics*[width=2.1in, height=2.1in, keepaspectratio=false]{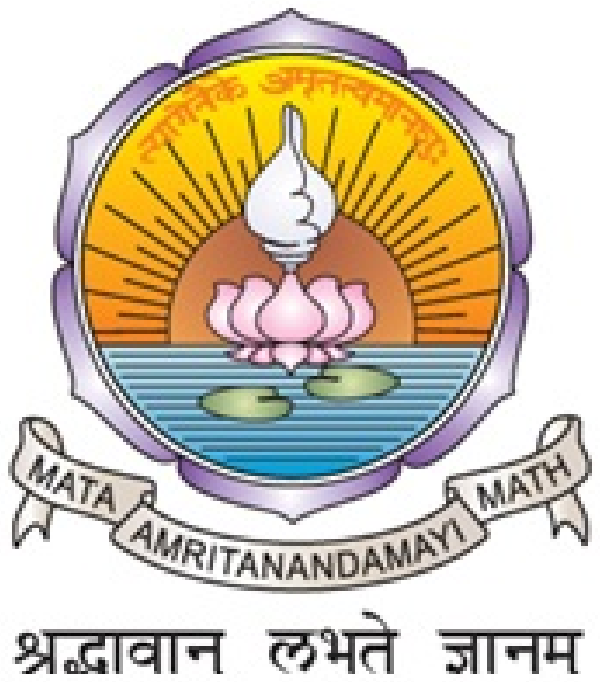}\\

\end{figure}

\textbf{\small{ Centre for Excellence in Computational Engineering and Networking}}\\

\vspace*{.2cm}

\large {\textbf{AMRITA SCHOOL OF ENGINEERING}}\\

\vspace*{.2cm}

\Large{ \textbf{AMRITA VISHWA VIDYAPEETHAM}}\\

\vspace*{.2cm}

\small{COIMBATORE} - 641 112 (\small{INDIA)}\\

\vspace*{.2cm}

\bf{\small{ JULY - 2015 }}

\end{center}

\end{large}

\baselineskip 1cm
\newtheorem{thm}{Theorem}[section]
\newtheorem{cor}{Corollary}[section]
\newtheorem{pro}{Proposition}[section]
\newtheorem{Lemma}{Lemma}[section]
\newtheorem{example}{Example}[section]
\newtheorem{rem}{Remark}[section]
\newtheorem{note}{Note}[section]
\newtheorem{defn}{Definition}[section]

\newpage

\pagenumbering{roman}

\begin{center}

\section*{Acknowledgement}

\end{center}

\baselineskip .7cm

{\normalsize  Let the \textbf{Almighty Lord}, be praised for his compassion, whose ample grace has helped me in the successful completion of my project.\\

I would like to extend my most sincere gratitude to all those who provided their assistance and co-operation during my project work.\\

\vspace*{.2cm}

First and foremost, I extend my sincere gratitude and heartfelt thanks to \textbf{Dr.K.P. SOMAN}, Head of the Department, Computational Engineering and Networking for giving me valuable suggestions and the freedom he has entrusted me throughout the course.\\

\vspace*{.2cm}

I also thank my guide \textbf{Dr. Anand Kumar M}, Assistant Professor (S.G), Computational Engineering and Networking for his dedicated and sincere efforts in introducing me to the latest topics in NLP and Machine Learning, which is invaluable and exceptional.\\

\vspace*{.2cm}

The \textbf{teaching} and \textbf{non-teaching} staffs of the Department of Computational Engineering and Networking has provided the necessities at various phases of the project and has offered timely help and cooperation upon request for which I am grateful. \\

\vspace*{.2cm}

My \textbf{family}, \textbf{friends} and \textbf{well-wishers} have always offered me moral and emotional support and encouragement, which helped me to strive forward throughout the course of study, which I am deeply indebted and grateful to.\\

\newpage
\clearpage
\tableofcontents
\addcontentsline{toc}{chapter}{\listfigurename}
\newpage
\listoffigures
\clearpage
\addcontentsline{toc}{chapter}{\listtablename}
\newpage
\listoftables
\clearpage

\newpage
\chapter*{Abstract\hfill} \addcontentsline{toc}{chapter}{Abstract}
Languages shared by people differ in different regions based on their accents, pronunciation and word usages. In this era sharing of language takes place mainly through social media and blogs. Every second swing of such a micro posts exist which induces the need of processing those micro posts, in-order to extract knowledge out of it. Knowledge extraction differs with respect to the application in which the research on cognitive science fed the necessities for the same. This work further moves forward such a research by extracting semantic information of streaming and batch data in applications like Named Entity Recognition and Author Profiling. In the case of Named Entity Recognition context of a single micro post has been utilized and context that lies in the pool of micro posts were utilized to identify the sociolect aspects of the author of those micro posts. In this work Conditional Random Field has been utilized to do the entity recognition and a novel approach has been proposed to find the sociolect aspects of the author (Gender, Age group).

\newpage
\baselineskip1cm
\pagenumbering{arabic}
\setcounter{page}{1}

\chapter{Introduction}
The amount of language sharing through internet is gains rapid growth due to use of the social media resources like Facebook, Twitter, LinkedIn, Pinterest and Instagram. It has surveyed that significant percent of people active in multiple social media resources \cite{1}. This positive growth ensures and encourages the internet marketing among users on that particular resource. It has been used business organization for the marketing, advertising and for connecting with customers. Hence the need of discovering knowledge from the shared language are entitled and put it into the research, which can help the business organization for observing their state among their customers, for understanding the products popularity among the users and for increasing the customers. Unlike natural languages the language shared on social media are tiny and unfortunate to extract information out of it \cite{2}. Named Entity Recognition is one among the knowledge discovery method coined on $19^{th}$ century \cite{3}, which can identify real world entities like PERSON, PLACE and ORGANIZATION from micro posts. This could be utilized as identifying the popularity of particular resource among users in a social network \cite{4}. Author profiling is an another application which plays major role in cognitive science. This is used for identification and prediction of sociolect aspects about the user in social networks \cite{5}. It involves business perspective applications like identifying products popularity among the users with respect to their personality, age group and gender, social network itself utilizes this to suggesting application to their users. Along with positive growth, malicious users also grows who are meant to extract the sensitive private information about unaware and unknown users \cite{6}. As per the norms by the social network resources user must not to have more than one account and users must not to be under age 13. It has been noted that 83.09 millions of fake accounts exist among 955 millions of monthly active users \cite{7}. Now a days auto generated accounts are utilized to give positive feed back about their own product in-order to increase their popularity. These ensures the need of author profiling in social media. The following chapter describes the works experimented and discusses about its outcome. 

\chapter{Distributional Hypothesis in Author Profiling}

\section{Introduction}
People started revolving around authorship tasks, right from the ancient Greek playwright times \cite{8}. Recognizing the age, gender, native language, personality \cite{9} and many more facets that frame the profile of a particular person, with respect to their writing style has become obligatory due to the plethora of available texts which gets multiplied every microsecond. It finds application in different zones like forensic security, literary research, marketing \cite{10}, industries, online messengers and chats in mobile applications, in medical applications to treat neuroticism and many more. Forensic Linguistics came into existence only after 1968. In this sector, police register is one of the area under security, in which the statements taken down by the police act as a source text for AP \cite{8}. Legal investigation continues its examination on all fields of suspicion. In marketing, online customer reviews in blogs and sites helps the consumers in deciding his/her choice about shopping a product \cite{11}. Detecting the age and gender of the person who posted his/her feedback paves way for the owners to improve their business strategy. Industries get benefited with emptier suggestions and reviews from which they could group the most likely products based on the gender and age.\\
Popular and the most often used social medias are Twitter and Facebook. Recent survey (first quarter of 2015) shows that every month there are about 236 million users who sign up to the micro blogging site-Twitter and 1.44 billion users to Facebook \cite{12}\cite{13}.With the increase in number of users the need to detect the AP has become an essential one. Based on the selection of specific applications, advertisements and groups follow Social Medias to discriminate the male and the female user with their age groups. There may also be anon who tend to have many fake id's and post messages and chat with innocent people in order to trap them. Katherine M. Hertlein and Katrina Ancheta have described such scenarios in their work on the usage of technology in relationships as a survey \cite{14}. John D. Burger et al came up with a model to discriminate gender on Twitter in which 213 million tweets were collected from 18.5 million users of various languages. They ended up with 92\% accuracy with the best classifier and 76\% accuracy for the classifier which relied on tweets \cite{15}.\\
In general Machine Learning (ML) algorithm can be used to attain the objective if subjected to relevant features. In existing methods the commonly used features for AP are author's style based features (punctuation marks, usage of capitals, POS tags, sentence length, repeated usage of words, quotations), content based features (topic related words, words present in dictionaries), content and typographical ease, words that express sentiments and emotions with emoticons, special words from which information could be extracted, collocations and n-grams \cite{16}\cite{17}. These features vary with respect to the topic, genre and language.
In ML, the low dimensional condensed vectors which exhibit a relation between the terms, documents and the profile was built using Concise Semantic Analysis (CSA) in order to create the second order attributes (SOA) and was classified using a linear model which became sensitive to high dimensionality problems \cite{18}. This system was further extended in 2014 to make it more precise in profiling. With the generation of highly informative attributes (creating sub profiles) using Expectation Maximization Clustering (EMC) algorithm the system built was able to group sub classes within a cluster and exhibit a relation between sub profiles. Though the system found success in both the years, still it was dependent to the language and genre \cite{19}.\\
The syntactic and lexical feature utilized in earlier models vary with respect to the morphological and agglutinative nature of the language. These features also varies with respect to domain in which AP was performed. There exists a conflict in classification algorithms to learn from these features in order to build a unified and effective classification model which is independent of domain and language. This can be observed from systems performance in PAN - AP shared task \cite{9}. In order to overcome these conflicts, this paper proposes a Life Time Learning Model based on distributional semantics. Distributional semantics paves way to advances in research in cognitive science by including statistical features of word distribution along with traditional semantic features utilized in Latent Semantic Analysis \cite{20}\cite{21}. It is clear that sexual aspects and vocabulary knowledge of a person varies due to human's cognitive phenomena which induces and also limits particular gender/age group people to utilize certain range of words to convey their message \cite{22}. Thus by utilizing this word distribution in word space, the gender and age group of a particular author is identified in this paper. The basic idea is utilizing the distribution of term's frequency in an author's document to aggregate the semantic information and promote a set of constraints for finding distributional hypothesis of that author's document. Along with AP this paper also proposes Life Time Learning Model (LTLM) by considering the limitation of constructing a VSM model and word space with commodity hardware machines. This is because when AP comes into real world applications like AP in social media the terms existence are tremendous, open class and elevates every now and then which has turned to a Big-Data problem \cite{23}. This paper encompasses a novel, principled word space based model for solving language independent AP on Big Data by utilizing the proposed Life Time Learning Algorithm.

\section{Related works}
John D Burger et al collected large number of tweets and also evaluated it with people work using Amazon Mechanical Turk (AMT). Their data included 213 million tweets on the whole from 18.5 million users. They started their data collection on April 2009 and completed it by 2011. Tweets collected were multilingual. As tweets include many more contents like emoticons, images etc., feature extraction part was limited to a particular n-gram length with total distinct features of 15,572,522. Word level and character level n-grams were chosen. There was no language specific processing done but instead only n-gram counts were taken into account. Once features were extracted classifiers namely SVM, Naive Bayes and Winnow2 were evaluated out of which Winnow2 performed exceptionally well with an overall accuracy of 92\%. Human performance was also cross checked and they found that only 5\% of the total users were able to classify tweets with good accuracy. Their work was done only for gender classifying gender information. Here other profiling informations were not taken into consideration \cite{15}.\\
Lizi Liao et al told that The entrance to colossal measure of client produced information empowers them to examine lifetime semantic variety of individuals. Their paper introduces an age theme show that mutually models inactive subjects and client's ages. The center reason of the model is that age impacts the point piece of a client, and every subject has an interesting age conveyance. Substance and age is consequently joined to shape the watched lexical varieties. Their model reveals cognizant subjects and their age appropriations, offering experiences into the ways of life of individuals of distinctive age bunches. Trials demonstrate that the model beats solid options during a time expectation assignment.They made use of Gibbs EM algorithm for evaluating their model. They were able to find information of both word distribution and age distribution from the sample of twitter data thay collected. They treated tweets as bag of words content thus performing well and effectively mapping the topic to ages \cite{23}.\\
A. Pastor López-Monroy and Manuel Montes-y-Gómez framed their methodology by utilizing the thought of second request properties (a low dimensional and thick record representation), yet goes past consolidating data among every objective profile. The proposed representation extended the examination fusing data among writings in the same profile, this is, they concentrated in sub-profiles. For this, they naturally discover sub-profiles and assemble report vectors that speak to more itemized connections of archives and sub-profile records. They contrast the proposed representation and the standard Bag-of-Terms and the best strategy in PAN13 utilizing the PAN 2014 corpora for AP undertaking. Results show proof of the helpfulness of intra-profile data to focus sex and age profiles. The sub profile or intra-profile information of each author was found using Expectation Maximization Clustering (EMC) algorithm. They were a participant at PAN 2014 and their results were placed among the first three results of the PAN AP shared task 2014 \cite{19}.\\
Suraj Maharjan, Prasha Shrestha, and Thamar Solorio have used MapReduce programming standard for most parts of their preparation process, which makes their framework quick. Their framework uses word n-grams including stop-words, accentuation and emoticons as components and TF-IDF (term recurrence reverse report recurrence) as the measuring plan. These were bolstered to the logistic relapse classifier that predicts the age and sexual orientation of the creators. Mapreduce distributed their tasks among many machines and made their work more easy and fast. This was their added advantage and their obtained results were in winning positions for both the testing and training results given by PAN \cite{10}.\\
Xiufang Xia compared the contrasts in the middle of men and ladies by utilizing dialect information. Their paper basically talks about the distinctions from the parts of articulation, sound, vocabulary, grammar, conduct, states of mind, and non-verbal contrasts in utilizing dialect in the middle of men and ladies. Other than the distinctions in different angles, their paper tries to record the progressions of these distinctions. On the premise of these distinctions and changes, their paper likewise tries to make some clarification to these distinctions and changes. Sexual orientation as a diagnostic classification keeps on spurring analysts in numerous regions. Their paper has seen the contrasts between the utilization of dialect of men and ladies from a few angles, and also made numerous distinctions in utilizing dialect between the two sexual orientations, furthermore there are a few progressions through time. They expected that with the advancement of society, will be less contrasts in the utilization of dialect. Dialect, as an instrument of human correspondence, will be enhancing step by step, and this needs the exertion of both men and ladies \cite{22}.\\
Feng Zhu and Xiaoquan (Michael) Zhang analyzed how item and buyer attributes direct the impact of online purchaser surveys on item deals utilizing information from the computer game industry. The discoveries demonstrate that online surveys are more persuasive for less prominent recreations and amusements whose players have more noteworthy Internet experience. Their work shows differential effect of shopper audits crosswise over items in the same item classification, and recommends that organization's internet promoting procedures ought to be dependent upon item and buyer qualities \cite{11}.\\

\section{Mathematical background}
\subsection{Problem Definition}
In general the solution is to build a training model from the given problem set and to map each document's author to specific gender and age group. The problem set $P$ consist of sub problems $\left \{ p_{1}, p_{2}, p_{3} ... \right \}$ which differs in language and genre. 
\subsection{Training Phase}
Step 1 - Constructing document - term matrix $\left [ V_{i,j} \right ]_{m\times n}$ \cite{17}, where $n$ is total number of documents in subset $p_{i}$, $m$ is size of the vocabulary and
\begin{equation}
\left[{V}_{i,j} \right]=term\, frequency\left({v}_{i,j} \right)\left(1 <i>n\right)\, and\, \left(1 <j>m\right)
\end{equation}
Step 2 - Under lying semantic information and relation between words can be obtained using latent vector by finding Eigen vectors of $\left[V{V}^{T} \right]$ which is column space of  $V$ \cite{24}. Thus the computed latent vectors spans the word space by satisfying the following condition,
\begin{equation}
Ax=\lambda xA=\left[V{V}^{T} \right],
\end{equation}
where $x$, $\lambda$ is Eigen vector and Eigen value of $A$. This literally means that projecting $A$'s rows on $x$ neither stretches or contracts the vector $x$, nor changes the direction which ensures that the originality of distribution of word remains unchanged. Thus the Eigen vectors of $A$ are computed by finding vectors from right null space of Eigen values shifted $A$. This can be represented as,
\begin{equation}
\left(A-{\lambda }_{k}I \right){x}_{k}=0\left(1<k>n \right)
\end{equation}
\begin{equation}
\left [ W \right ]_{n\times n} =\left [ x_{1},x_{2},...,x_{n} \right ]
\end{equation}
Step 3 - The statistical features of word distribution in word space are computed in order to build supervised classification model which is independent to domain and language \cite{26}. Statistical features include the marginal decision boundaries with respect to word distribution in each document vector $v_{i}$  based on each class which has to be classified. By taking $v_{i}$ as a random variable, along with mean and variance of a distribution, the asymmetry of distribution of words, peakedness of distribution of words are also taken into account. This is expressed as,
\begin{equation}
\left [ F \right ]_{n\times s}=statistics\left ( \left [ W \right ]_{n\times n}\right )
\end{equation}
Where, $s$ is number of statistical features and $F$ is feature matrix for building classification model. From the above it is clear that the extracted features are only dependent on author's word distribution in a document.\\
Step 4 - To limit the number of feature vectors which contributes less towards classification the test of goodness fit is carried out. The minimum probability of $F_{i}$ coming from same distribution and maximum probability of  $F_{i}$  coming from opposite probability gets eliminated by performing Kolmogorov-Smirnov test between entropy distribution vector in a class and feature vectors in all class \cite{27}. This is expressed as,
\begin{equation}
C=\max_{a}\left ( \left | F_{i}\left ( a \right ) -\sum_{c}F\left ( a \right ) \right | \right )
\end{equation}
The sum of all the feature vectors which belongs to a particular class is computed, which is called as entropy distribution of that class. By utilizing this entropy distribution of a class, probabilities of each feature coming out from similar and opposite classes are computed. In each case the worst 5\% of features are eliminated.\\
Step 5 - In order to build classification model, the regression relation between the feature and the respective class are constructed using Random Forest tree algorithm \cite{28} which is a collection of Decision trees that formulates the classification rule based on randomly selected features in training set \cite{29}. From $L=\left \{ \left ( y_{i},F_{i} \right ),1<i>n \right \}$ the subset of $L_{b}$ is formed using $\sqrt{s}$ and $b$ number of aggregate predictor is built. Then final predictor is built by,
\begin{equation}
\varphi_{b}\left ( F \right )=max_{J}H_{J}
\end{equation}
Where, $J$ is number of decision trees and   $H_{J}=\left \{ \varphi \left ( F,L_{b} \right ) = J\right \}$.\\
The gender and age group classification model is built using hierarchical method. In training there are two models built for gender and age classification. This is explained in testing phase.\\
\subsection{Testing Phase}
Step 6 - As similar to training set except Step 4 and Step 5 the test set $p_{t}=\left \{ d_{1},d_{2},...,d_{n1} \right \}$ follows all remaining steps to compute feature vectors. Further classification of document into gender and age group is performed using $b$ aggregate predictors in hierarchical method. The final class is assigned based on equation aggregate. The test features $\left [F_{t} \right]$ are initially classified into male/female and padding is done as an additional feature for further age group classification.  \\ 
The algorithm for training and testing are shown below,\\
\begin{algorithm}[!h]
\caption{Training}
Input $p_{i}=\left \{ {d_{1},d_{2},...,d_{n}} \right \}$ \\
 \For { i=1 to n }
{
 $\left [ V \right ]=VSM\left ( d_{i} \right )$ \\
 }
 $\left [ W \right ]=eigen\, vector\left ( \left [ VV^{T} \right ] \right )$\\
 $[F]=statistical feature([W])$\\
 ${[F_{final}]=feature\, elimination\left ( \left [ F \right ] \right )}$\\
 ${model_{gen}=rft\left ( \left [ F_{final},b \right ] \right )}$\\
 ${model_{age}=rft\left ( \left [ F_{final},gender \right ] \right )}$\\
\end{algorithm}
\begin{algorithm}[!h]
\caption{Testing}
Input $p_{t}=\left \{ {d_{1},d_{2},...,d_{n}} \right \}$ \\
 \For { i=1 to n }
{
 $\left [ V \right ]=VSM\left ( d_{i} \right )$ \\
 }
 $\left [ W \right ]=eigen\, vector\left ( \left [ VV^{T} \right ] \right )$\\
 $[F]=statistical feature([W])$\\
 ${[F_{final}]=feature\, elimination\left ( \left [ F \right ] \right )}$\\
 ${y_{gen}=predict\left (model_{gen},\left [F \right ] \right )}$\\
 ${y_{age}=predict\left (model_{age}, \left [F, y_{gen}\right ] \right )}$\\
\end{algorithm}
\subsection{Life Time Learning Model (LTLM)}
As mentioned earlier since the approach is independent of language and domain it allows batch processing to construct the classification model. While applying traditional algorithm in commodity machines it faces the problem (Out of Memory) of handling high dimensional space. The interesting thing is that by pairing LTLM with previously stated algorithm, the proposed algorithm can handle the Big-Data and also paves way to train a model with new features when ever required. The given dataset is divided into subsets, with subsets of approximately equal sizes and classes. This ensures the space in which the feature extracted are of same dimension. Then the training model is constructed after forming feature matrix from all these batches.  \\
\subsubsection{LTLM algorithm}
\begin{algorithm}[!h]
\caption{LTLM}
$p_{t}={ d_{1},d_{2},...,d_{n}}$ \\
$p^{'}_{j}=make subsets(p_{i})$ \\
 \For { j=1 to N }
{
 $p^{'}_{j}=d_{1},d_{2},...,d_{n}$ \\
\For{i=1 to n}
{
$[V]=VSM(d_{i})$\\
}
$[W]=eigen vector ([ VV^{T}])$\\
$[F]=statistical feature([W])$\\
${[F_{1}]=padding (F1,F)}$\\
}
${F_{final}=feature elimination (F1)}$\\
$model_{gen}=rft ([ F_{final}],b)$\\
$model_{age}=rft ([ F_{final},gender])$\\
\end{algorithm}
The algorithm is designed in such a way that it can limit the number of feature to construct training model. This is because when more and more features are added building the training model itself becomes complex and there exists a chance of irrelevant features getting added. Such problems can be solved and numbers of features could be limited by utilizing feature elimination.
\section{Experimental works and Results}
\subsection{Data-set}
Finding human sociolect aspect is current demanding research field, which includes many more social and business applications. To motivate and to bring new ideas many universities has started conducting shared tasks such as author profiling at PAN CLEF workshop, Native language identification at BEA workshop and Personality recognition at ICWSM. The data-set chosen for this experimentation was taken from the PAN CLEF AP 2013 and 2014 workshop which are built with challenges involved in real word applications. The 2013 corpus incorporates blogs browsed from the social networking site Netlog (even-handed in gender and unstable by age group) and to make it more realistic texts of sexual predators and adult-adult conversations were also included of two languages (English and Spain). The 2014 data-set comprises of different genres with fine grained categorization in age groups. As a part of PAN AP 2013 corpus, blogs were extracted from LinkedIn profiles, Twitter data from Replab 2013 corpus and hotel reviews extracted from the site TripAdvisor. More statistics about data-set is given in following table.
\begin{table*}[!ht]
\centering
\caption{PAN Author Profiling 2013 Training set Statistics}
\scalebox{0.7}{
\begin{tabular}{|c|cccccc|}
\hline
 \multicolumn{1}{c}{} & & & \\ [\dimexpr-\normalbaselineskip-\arrayrulewidth]
 \textbf{Language} & \multicolumn{3}{c|}{ \textbf{\#Docs per Male/Female}} & \multicolumn{1}{c|}{ \textbf{\# Docs per }} &\multicolumn{1}{c|}{ \textbf{\#Words Before }} & \multicolumn{1}{c|}{ \textbf{\# Words After }} \\ 
 & \multicolumn{3}{c|}{ \textbf{Age Group }} & \multicolumn{1}{c|}{} & \multicolumn{1}{c|}{} & \multicolumn{1}{c|}{} \\ \cline{2-4}
 & \multicolumn{1}{c|}{\textbf{10s}} &\multicolumn{1}{c|}{\textbf{20s}}  &\multicolumn{1}{c|}{\textbf{30s}} & \multicolumn{1}{c|}{\textbf{Male/Female}}  & \multicolumn{1}{c|} {\textbf{Cleaning}} & \multicolumn{1}{c|} {\textbf{Cleaning}}\\[0.5ex]\hline
{English}& \multicolumn{1}{c|}{8600} & \multicolumn{1}{c|}{42900} & \multicolumn{1}{c|}{66800} & \multicolumn{1}{c|}{118300} & \multicolumn{1}{c|}{14-61161} & \multicolumn{1}{c|}{4-61040} \\[0.5ex]\hline
{Spanish}& \multicolumn{1}{c|}{1250} & \multicolumn{1}{c|}{21300} & \multicolumn{1}{c|}{15400} & \multicolumn{1}{c|}{37950} & \multicolumn{1}{c|}{14-12919} & \multicolumn{1}{c|}{4-11808} \\[0.5ex]\hline   
 \end{tabular}}
 \end{table*}

\begin{table*}[!ht]
\centering
\caption{PAN Author Profiling 2013 Test set - 1 Statistics}
\scalebox{0.7}{
\begin{tabular}{|c|cccccc|}
\hline
 \multicolumn{1}{c}{} & & & \\ [\dimexpr-\normalbaselineskip-\arrayrulewidth]
 \textbf{Language} & \multicolumn{3}{c|}{ \textbf{\#Docs per Male/Female}} & \multicolumn{1}{c|}{ \textbf{\# Docs per }} &\multicolumn{1}{c|}{ \textbf{\#Words Before }} & \multicolumn{1}{c|}{ \textbf{\# Words After }} \\ 
 & \multicolumn{3}{c|}{ \textbf{Age Group }} & \multicolumn{1}{c|}{} & \multicolumn{1}{c|}{} & \multicolumn{1}{c|}{} \\ \cline{2-4}
 & \multicolumn{1}{c|}{\textbf{10s}} &\multicolumn{1}{c|}{\textbf{20s}}  &\multicolumn{1}{c|}{\textbf{30s}} & \multicolumn{1}{c|}{\textbf{Male/Female}}  & \multicolumn{1}{c|} {\textbf{Cleaning}} & \multicolumn{1}{c|} {\textbf{Cleaning}}\\[0.5ex]\hline
{English}& \multicolumn{1}{c|}{740} & \multicolumn{1}{c|}{3840} & \multicolumn{1}{c|}{6020} & \multicolumn{1}{c|}{12720} & \multicolumn{1}{c|}{14-25110} & \multicolumn{1}{c|}{4-24862} \\[0.5ex]\hline
{Spanish}& \multicolumn{1}{c|}{120} & \multicolumn{1}{c|}{1920} & \multicolumn{1}{c|}{1360} & \multicolumn{1}{c|}{3400} & \multicolumn{1}{c|}{14-8171} & \multicolumn{1}{c|}{4-7884} \\[0.5ex]\hline   
 \end{tabular}}
 \end{table*}

\begin{table*}[!ht]
\centering
\caption{PAN Author Profiling 2013 Test set - 2 Statistics}
\scalebox{0.7}{
\begin{tabular}{|c|cccccc|}
\hline
 \multicolumn{1}{c}{} & & & \\ [\dimexpr-\normalbaselineskip-\arrayrulewidth]
 \textbf{Language} & \multicolumn{3}{c|}{ \textbf{\#Docs per Male/Female}} & \multicolumn{1}{c|}{ \textbf{\# Docs per }} &\multicolumn{1}{c|}{ \textbf{\#Words Before }} & \multicolumn{1}{c|}{ \textbf{\# Words After }} \\ 
 & \multicolumn{3}{c|}{ \textbf{Age Group }} & \multicolumn{1}{c|}{} & \multicolumn{1}{c|}{} & \multicolumn{1}{c|}{} \\ \cline{2-4}
 & \multicolumn{1}{c|}{\textbf{10s}} &\multicolumn{1}{c|}{\textbf{20s}}  &\multicolumn{1}{c|}{\textbf{30s}} & \multicolumn{1}{c|}{\textbf{Male/Female}}  & \multicolumn{1}{c|} {\textbf{Cleaning}} & \multicolumn{1}{c|} {\textbf{Cleaning}}\\[0.5ex]\hline
{English}& \multicolumn{1}{c|}{740} & \multicolumn{1}{c|}{3840} & \multicolumn{1}{c|}{6020} & \multicolumn{1}{c|}{12720} & \multicolumn{1}{c|}{14-11227} & \multicolumn{1}{c|}{4-24862} \\[0.5ex]\hline
{Spanish}& \multicolumn{1}{c|}{120} & \multicolumn{1}{c|}{1920} & \multicolumn{1}{c|}{1360} & \multicolumn{1}{c|}{3400} & \multicolumn{1}{c|}{14 - 8171} & \multicolumn{1}{c|}{4-7884} \\[0.5ex]\hline   
 \end{tabular}}
 \end{table*}

\begin{table*}[!ht]
\centering
\caption{PAN Author Profiling 2014 Data-set Statistics}
\scalebox{0.7}{
\begin{tabular}{|c|ccccccccc|}
\hline
 \multicolumn{1}{c}{} & & & \\ [\dimexpr-\normalbaselineskip-\arrayrulewidth]
 \textbf{Language} & \multicolumn{5}{c|}{ \textbf{\#Docs per Male/Female Age Group }} & \multicolumn{1}{c|}{ \textbf{\# Docs per }}& \multicolumn{1}{c|}{ \textbf{Total }} &\multicolumn{1}{c|}{ \textbf{\#Words Before }} & \multicolumn{1}{c|}{ \textbf{\# Words After }} \\ \cline{2-6}
\textbf{\& Genre}& \multicolumn{1}{c|}{\textbf{18-24}} &\multicolumn{1}{c|}{\textbf{25-34}}  &\multicolumn{1}{c|}{\textbf{35-49}}  & \multicolumn{1}{c|}{\textbf{50-64}} & \multicolumn{1}{c|}{\textbf{64-XX}} & \multicolumn{1}{c|}{\textbf{Male/Female}} &\multicolumn{1}{c|}{\textbf{\#Docs}} & \multicolumn{1}{c|} {\textbf{Cleaning}} & \multicolumn{1}{c|} {\textbf{Cleaning}}\\[0.5ex]\hline
{EN-Blogs}& \multicolumn{1}{c|}{3} & \multicolumn{1}{c|}{30} & \multicolumn{1}{c|}{27} & \multicolumn{1}{c|}{12} & \multicolumn{1}{c|}{2} & \multicolumn{1}{c|}{74} & \multicolumn{1}{c|}{147} & \multicolumn{1}{c|}{168-50124} & \multicolumn{1}{c|}{147} \\[0.5ex]\hline
{EN-Review}& \multicolumn{1}{c|}{180} & \multicolumn{1}{c|}{500} & \multicolumn{1}{c|}{500} & \multicolumn{1}{c|}{500} & \multicolumn{1}{c|}{400} & \multicolumn{1}{c|}{2080} & \multicolumn{1}{c|}{4160} & \multicolumn{1}{c|}{18-5908} & \multicolumn{1}{c|}{5-5879} \\[0.5ex]\hline
{EN-Twitter} & \multicolumn{1}{c|}{10} & \multicolumn{1}{c|}{44} & \multicolumn{1}{c|}{65} & \multicolumn{1}{c|}{30} & \multicolumn{1}{c|}{4} & \multicolumn{1}{c|}{153} & \multicolumn{1}{c|}{306} & \multicolumn{1}{c|}{328-54224} & \multicolumn{1}{c|}{11-7988} \\[0.5ex]\hline
{SP-Blogs}& \multicolumn{1}{c|}{2} & \multicolumn{1}{c|}{13} & \multicolumn{1}{c|}{21} & \multicolumn{1}{c|}{6} & \multicolumn{1}{c|}{2} & \multicolumn{1}{c|}{44} & \multicolumn{1}{c|}{88} & \multicolumn{1}{c|}{250-73764} & \multicolumn{1}{c|}{129-48976} \\[0.5ex]\hline
{SP-Twitter}& \multicolumn{1}{c|}{6} & \multicolumn{1}{c|}{21} & \multicolumn{1}{c|}{43} & \multicolumn{1}{c|}{16} & \multicolumn{1}{c|}{3} & \multicolumn{1}{c|}{89} & \multicolumn{1}{c|}{178} & \multicolumn{1}{c|}{114-32697} & \multicolumn{1}{c|}{3-3511} \\\hline        
 \end{tabular}}
 \end{table*}
\subsection{Experiment}
As detailed in problem definition section, the total documents are divided into 50 batches such that each have 4732 user documents in English and 1518 user documents in Spanish. In order to avoid the biasing problem the distribution of gender and age group are maintained almost equal in all the batches. As given in previous section the sub feature matrix is constructed from the all batches and computed features are concatenated to form single feature matrix which has size of $2,36,600 \times 9$ for English and $75900 \times 9$ for Spanish. This final matrix is utilized to construct the classification model which can built using soft-max classifiers. Here the classification model is built based on 100 decision trees, constructed to form the random forest tree. Primarily gender classification was performed and by feeding its result to feature matrix, age group was then classified. Since the algorithm proposed as language and genre independent model, by excluding feature selection from the algorithm, it is applied on 2014 data-set, which includes multiple genre and language.
\subsection{Results and Observations}
10 fold cross validation was performed and F1 measure calculated based on recall and precision of individual batches and was then put together to form the concatenated feature matrix of batches. It is observed that the F1 measure of individual batches varies $\pm 3$ from 80\% in gender classification and $\pm 2$ from 89\% in age group classification for English. Similarly $\pm 4$ from 79\% in gender classification and $\pm 3$ from 87\% in age group classification achieved for Spanish. As concatenated feature matrix grows the F1 measure of English became constant nearing 84\% in gender classification and 91\% in age group classification which varies in decimal point value. Similarly for Spanish it resulted in 81\% in gender classification and 86\% in age group classification. Up-to 22 batches it remains constant nearing 84\% for gender and 91\% for age group and beyond that concatenation to feature matrix improves accuracy by 2\% and 1\% respectively from previous F1 measures of English. For Spanish up-to 19 batches it resulted in 81\% in gender and 88\% in age group beyond that it improves accuracy by 1.5\% and 2\% with respect to constant previous measure, which also varies only in decimal values. The above performance was analyzed to observe the accuracy with the growing feature matrix. After computing the whole feature matrix from 50 batches, two classifier models were developed for gender and age group classification. The final F1 measure was calculated for test set 1 and test set 2 and also 10 fold cross validation performed on 2014 data-set. These are shown in the following tables and figures.

\begin{figure}[!h]\centering
\includegraphics[scale=0.75]{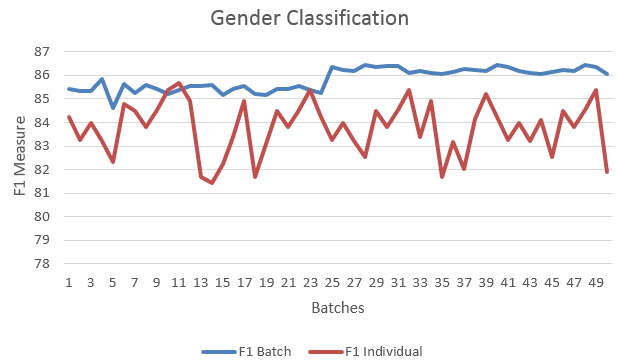}
\caption {10 Cross Validation of Training Batches - English Gender Classification \label{eng1}}
\end{figure}

\begin{figure}[!h]\centering
\includegraphics[scale=0.75]{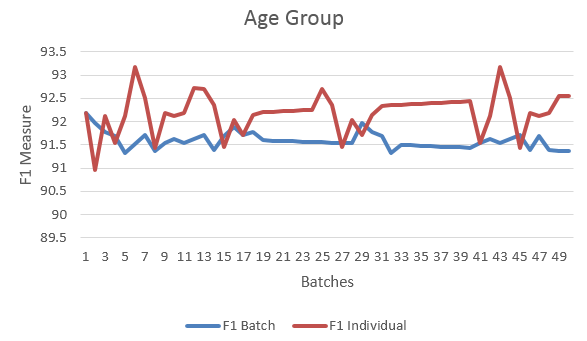}
\caption {10 Cross Validation of Training Batches - English Age Group \label{eng2}}
\end{figure}

\begin{figure}[!h]\centering
\includegraphics[scale=0.75]{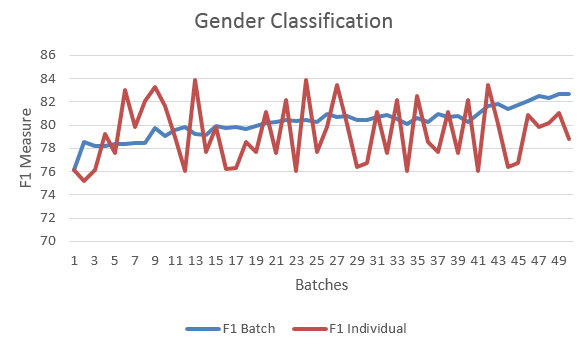}
\caption {10 Cross Validation of Training Batches - Spanish Gender Classification \label{eng3}}
\end{figure}

\begin{figure}[!h]\centering
\includegraphics[scale=0.75]{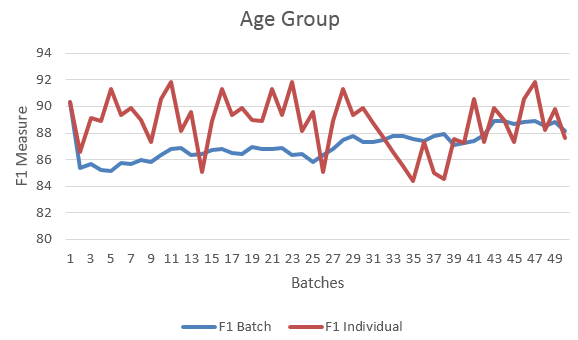}
\caption {10 Cross Validation of Training Batches - Spanish Age Group \label{eng4}}
\end{figure}

\begin{table}[!ht]
\centering
\caption{Results for 2013 PAN Author Profiling Data-set}
\begin{tabular}{|c|c|c|}
\hline
\multicolumn{1}{|c|}{\textbf{Language}} & \multicolumn{1}{|c|}{ \textbf{Gender F1 }} & \multicolumn{1}{|c|}{ \textbf{Age Group F1}}\\ \hline
\multicolumn{1}{|c|}{\textbf{English 1}} & \multicolumn{1}{c|}{81.32} & \multicolumn{1}{c|}{83.43}\\ \hline
\multicolumn{1}{|c|}{\textbf{Spanish 1}} & \multicolumn{1}{c|}{78.63} & \multicolumn{1}{c|}{81.21}\\ \hline
\multicolumn{1}{|c|}{\textbf{English 2}} & \multicolumn{1}{c|}{80.67} & \multicolumn{1}{c|}{81.12}\\ \hline
\multicolumn{1}{|c|}{\textbf{Spanish 2}} & \multicolumn{1}{c|}{78.63} & \multicolumn{1}{c|}{81.21}\\ \hline
 \end{tabular}
 \end{table}

\begin{table}[!ht]
\centering
\caption{Results for 2014 PAN Author Profiling Data-set}

\begin{tabular}{|c|c|c|}
\hline
\multicolumn{1}{|c|}{\textbf{Language \& Genre }} & \multicolumn{1}{|c|}{ \textbf{Gender F1 }} & \multicolumn{1}{|c|}{ \textbf{Age Group F1}}\\ \hline
\multicolumn{1}{|c|}{\textbf{EN-Blogs}} & \multicolumn{1}{c|}{80.75} & \multicolumn{1}{c|}{84.78}\\ \hline
\multicolumn{1}{|c|}{\textbf{EN-Review}} & \multicolumn{1}{c|}{75.45} & \multicolumn{1}{c|}{93.19}\\ \hline
\multicolumn{1}{|c|}{\textbf{EN-Twitter}} & \multicolumn{1}{c|}{78.99} & \multicolumn{1}{c|}{92.30}\\ \hline
\multicolumn{1}{|c|}{\textbf{SP-Blogs}} & \multicolumn{1}{c|}{72.93} & \multicolumn{1}{c|}{79.16}\\ \hline
\multicolumn{1}{|c|}{\textbf{SP-Twitter}} & \multicolumn{1}{c|}{85.71} & \multicolumn{1}{c|}{91.17}\\ \hline
 \end{tabular}
 \end{table}

\section{Conclusion and Future work}
With the global need for author profiling system, this experimentation has brought forth a simple and reliable model for finding the demographic features of an individual by extracting distributional semantics using LTLM algorithm. Achieved accuracy beats the systems which were submitted for 2013 and 2014 PAN author profiling shared task. Proposed model proved and acts as a global model which is independent of language, topic and genre. The proposed LTLM system has nearing 40\% greater accuracy than the 2013 top ranked system and nearing 20\% greater than the 2014 top ranked system. Even though the model produces greater accuracy, it takes several hours to extract feature from a single batch document. This can be avoided by implementing the same on a parallel or distributed computation frame work.

\chapter{Sequence Modeling for extracting Semantic Information of the Streaming Data}
\section{Introduction}
Language is the essential aspect in human communication. Natural Language Processing (NLP) deals with introducing the human-machine communication and develops the model that achieves the human performance in hearing, speaking, reading, writing and understanding. NLP applications require independent decisions by the system with the help of the prior knowledge given while training with the corpus. Named Entity Recognition (NER) is one such application. It refers to the identification of named entities from the text. Sequence modeling became mandatory when the knowledge extraction from the social media comes to picture. A short text gets updated every now and then. With the global upswing of such micro posts, the need to retrieve information from them also seems to be incumbent. This chapter focuses on the knowledge extraction from the micro posts by having entity as evidence and also the extracted entities are then linked to their relevant DBpedia source by featurization, Part Of Speech (POS) tagging, Named Entity Recognition (NER) and Word Sense Disambiguation (WSD). This chapter encompasses its contribution to Micropost'15 - Named Entity rEcognition Linking and Forum of Information Retrieval and Evaluation (FIRE 2014) tasks by experimenting existing Machine Learning (ML) algorithms. Micro posts has a pool of knowledge with scope in business analytics, public consensus, opinion mining, sentimental analysis and author profiling and thus indispensable for Natural Language Processing (NLP) researchers. People use short forms and special symbols for easily conveying their message due to the limited size of micro posts which has eventually built complexity for traditional NLP tools \cite{65}. Though there are number of tools, most of them rely on least ML algorithms which are effective for long texts than short texts. Thus by providing sufficient features to these algorithms the objective can be achieved. This chapter includes the NEEL and FIRE tasks carried out with the available NLP tools to evaluate their effect on entity recognition by providing special features available in tweets.  For example consider a sentence: \emph{Ramu joined Amrita as a Post Graduate student in Coimbatore on 7th August 2014.} Here 'Ramu' is a PERSON entity, 'Amrita' is an ORGANIZATION entity, 'Coimbatore' is a LOCATION entity and '7th August 2014' is DATE entity. NER is in need of external knowledge and non-local features in order to find the multiple occurrences of named entities in the text \cite{30}. Many fundamental decisions such as identifying the sequential named entities and noun chunks are made \cite{30}. Mostly the sequential named entities are chunks. NER plays a key role in NLP applications like machine translation, question-answering, information retrieval and extraction, and text summarization \cite{31}. NER is a subtask in Information Extraction which extracts the pre-specified types of information such as events, entities or relationships \cite{31}. NER systems are implemented using rule based approach and machine learning techniques on both supervised and unsupervised approach.  The commonly used machine learning approaches are Decision trees, Hidden Markov Model (HMM), Maximum Entropy Markov Model (MEMM), Conditional Random Fields (CRF), Support Vector Machine (SVM) etc \cite{32}. Many methods are implemented for Indian Languages and conventional approach is rule based approach that uses gazetteer or dictionary and patterns of named entities \cite{32}. NER is also used in various domain focused applications such as genetics, tourism etc. The tourism domain for NER focuses on named entities such as identifying place, water bodies like rivers, lakes etc. It also includes entities like monuments, museums, religious places and parks. The NER system for genetics involves identification of DNA names, protein names and gene names etc. This task focuses on Embedded Named Entity Recognition. It is also referred to as nested NER which means named entity having other named entities inside it. Nested NER is mostly used in case of Noun Chunks. The beginning of the chunk is tagged as 'B-tag' and middle portion of the chunk is tagged as 'I-tag' and outside the chunk is tagged as 'O-tag'. This kind of tagging is BIO-tagging. For example consider the sentence: \emph{Seeta was in Government of India for 3 to 5 years.} Here 'Seeta' is B-PERSON, 'Government of India' is a chunk of three words where 'Government' is B-GOV, 'of' is I-GOV and 'India' is I-GOV. In this Noun chunk we have word 'India' which alone a named entity. So, this token is tagged in second level as B-NATION. The training corpus for this task consists of three levels of named entity tag. Consider the example from training data: \emph{He is also primarily a Chicago Bears football fan in the NFL.} The tags for this sentence are in Table \ref{NETags}.
\begin{table}
\centering
\caption{Three levels of Named Entity Tags\label{NETags}}
\begin{tabular}{|c|c|c|c|} \hline
Word & Level-1 & Level-2 & Level-3 \\ \hline
Chicago & B-Person & B-Association	& B-Location \\
Bears & I-Person & I-Association & B-Nonhuman \\
Football & I-Person & I-Association & B-Sports \\
Fan & I-Person & O & O \\
\hline\end{tabular}
\end{table}
\section{Named Entity Recognition}
Unlike English, Indian languages face many challenges in NER due to its morphological rich nature. Also, it is a hard task since we have unbounded number of possible names \cite{32}. English language has a capitalization feature which helps in identifying the entity whereas it is not found in Indian languages. Indian Language faces various issues and challenges such as Variation in Usage, Ambiguity and agglutination. For example Tamil words contain various features like PNG markers, post-positions and case markers which help in entity types identification. Rule based approach is insufficient for Tamil since the position or ordering of words has no strong bearing whether the word is named entity or not \cite{33}.

\subsection{Related works}
Named Entity Recognition is one of the leading research work done in Natural Language Processing. NER can be implemented using various approaches like Rule based approach and Machine Learning based approach such as supervised, unsupervised and semi-supervised. Rule based approach require language expertise and it is very time consuming whereas machine learning approach does not require language engineers and the system learns based on the large amount of training data. Many systems for NER are developed using supervised algorithms such as Maximum Entropy Model, Hidden Markov Model, CRF and SVM. These statistical models also makes use of gazetteer in order to improve the performance of the system \cite{34}. Semi-supervised and unsupervised machine learning approach are also used for many languages.

Ralph Grishman worked on rule based approach by using dictionary for English language consisting of name of all countries and major cities \cite{35}. A named entity recognition system for English language using machine learning approach of maximum entropy algorithm was done by Andrew Borthwick \cite{36}. A unsupervised NER system for English was implemented by David et al and compared the results with supervised algorithm. It uses classical named entity tags such as Person, Location and Organization and evaluated the corpus with mentioned tags and other named entity tags also like car brands \cite{37}. A paper titled Focused Named Entity Recognition using Machine Learning by Li Zhang, Yue Pan and Tong Zhang described about focused named entities which is useful for document summarization. This uses NER as a classification problem and a comparison is done for various classification algorithms \cite{38}. A survey on NER for Indian languages are done by Pallavi et al in which they mentioned about the NER system developed for various Indian languages and their approaches \cite{39}. An another survey on Indian languages were done by Prakash and Shambhavi mentioning about the approaches that was used for NER system \cite{40}. Named Entity Recognition systems for Indian languages are designed using Maximum Entropy model. This uses the gazetteer information and Language specific rules \cite{41}. A supervised and unsupervised system is developed by Ekbal and Sivaji which is language independent. They used annotated corpora for both Bengali and Hindi with twelve NE classes \cite{42}. Darvinder and Vishal made a survey on the work done in NER for English and Indian languages. They mainly concentrated on Punjabi language and NER system developed on Punjabi language \cite{34}. NER was also implemented in other languages such as Arabic using integrated machine learning techniques \cite{43}. NER system is also developed for Manipuri language which is a less computerized Indian language \cite{44}. This was performed by Thoudam Doren et al using Support Vector Machines \cite{44}. This resulted in accuracy of 94.59\% as F-Score. Asif Ekbal and Sivaji Bandyopadhyay also developed a NER system for Bengali using SVM that makes use of different context information \cite{45}.

A domain focused application on nested NER was done by Vijayakrishna and S. L. Devi in Tamil language using Conditional Random Fields \cite{32}. An automated system for NER was also developed in Tamil Language using hybrid approach by Jeyashenbagavalli, Srinivasagan and Suganthi \cite{46}. Malarkodi et al described the various challenges faced during automatic NER system development using CRF for Tamil language \cite{47}. Features are more important in order to improve the accuracy of the system. Yassine Benajiba, Mona Diab and Paolo Rosso investigated different sets of features into CRF and SVM \cite{48}. They described the impact of each feature with the named entity and combined to find optimal machine leaning approach \cite{48}. Surya Bahadur Bam, Tej Bahadur Shahi analyzed the accuracy of their system which uses SVM algorithm with 3 different sizes of training corpus of Nepali text. They use a set of features and a small dictionary to analyze its performance \cite{49}.

\subsection{Mathematical Background}
This work focuses on using Conditional Random Field (CRF) for Named Entity Recognition (NER). CRF is implemented in CRF toolkit for tagging sequential data. CRF toolkit is a customizable tool in which feature sets can be redefined and fast training is possible \cite{50}.  The mathematical formulations for the approach is described below:
\subsection{Conditional Random Fields}
Conditional Random Field (CRF) is a probabilistic framework which is used for segmenting and labelling the structured data. CRF has various advantages over Hidden Markov Model and Maximum Entropy Markov Models. It outperforms both the approaches for various NLP tasks.

Let $G=(V,E)$ be a graph to extend that $Y=(Y_{v\epsilon V})$, so $Y$ is indexed by the vertices of $G$. Then $(X,Y)$ is a CRF in case, when conditioned on $X$, the random variables $Y_{v}$ obey the Markov property \cite{51} \cite{52}.
\begin{equation}
p(Y_{v}|X,Y_{W},w\neq v) = p(Y_{V}|X,Y_{W},w\sim v)
\end{equation}
where that $W$ and $V$ are neighbours in $G$, which means on the observation $X$ a random field is globally conditioned and assuming the graph $G$ is fixed by having edges as a features $(e)$ and $(v)$ as tags which are considered as its cliques. Word sequence taken as $X=(X_{1}, X_{2},...,X_{n})$ and $Y=(Y_{1}, Y_{2},...,Y_{n})$ is the tag sequence.
The distribution over the tag sequence $Y$ given $X$ word sequence has the form and shows it has single exponential model by considering the global tags given the word sequence \cite{51},
\begin{equation}
p_{\theta}(y|x) \;\; \alpha \;\; exp(\sum _{e\epsilon E,k} \lambda _{k} f_{k}(e,y|_{e},x)+\sum _{v\epsilon V,k} \mu _{k} g_{k}(v,y|_{v},x))
\end{equation}
where, $k$ is the length of the sequence, $\mu _{y,x}$ is the parameter for tag-word pair and $\lambda _{y,y^{'}}$ is the parameter for tag-tag pair in the training set. This plays a similar role to the usual Hidden Markov Model parameters state pair $p(y^{'},y)$ and state observation pair $p(x|y)$. Similarly $f_{k}$ and $g_{k}$ are features in the training set.

The parameters $(\theta = (\lambda _{1}, \lambda _{2},...;\mu _{1}, \mu _{2},...))$ are determined from training data $D={(x^{(i)},y^{(i)})}_{i=1}^{N}$ with empirical distribution $\tilde{p}(x,y)$. The normalized function $Z(X)$ for conditional probability is dependent on the observation.

\begin{equation}
p_{\theta}(y|x) = \frac{exp(\sum _{e\epsilon E,k} \lambda _{k} f_{k}(e,y|_{e},x)+\sum _{v\epsilon V,k} \mu _{k} g_{k}(v,y|_{v},x))}{Z_{\theta}(X)}
\end{equation}

In order to normalize, the function has to compute $Z_{\theta}(X)$ but is not an easiest task since its need to compute sum of the numerator over all possible tags available in the training set. The above requirement can be expressed as bunch of nested summation.
\begin{equation}
Z_(X)=\sum _{y_{1}^{'}} \sum _{y_{2}^{'}}...\sum _{y_{k}^{'}}exp \;\; \tilde{p}(x,y)
\end{equation}
The computation of $Z(X)$ is heavy because of nested summations. So the overall summation is done by number of partial summation.

\section{Experiments and Observations}
\subsection{FIRE - 2014}
The data-sets used for this work is obtained from Forum of Information Retrieval and Evaluation (FIRE 2014) NER track which are collected from Wikipedia and other resources such as blogs and online discussion forums. Experiments include training, testing and evaluating the named entity tags using CRF and SVM. In the process of system development, we trained the model using machine learning approaches like CRF and SVM. The obtained model is tuned with the help of development data. This validated model obtained from learning is used for tagging the test data. The size of training, development and test data-set for all four languages used in the system is given in Table \ref{Size_of_Dataset}. The number of sentences and words used in training and testing for all four languages are described in Table \ref{leng}. The layout for training and testing corpus are discussed in following sections.
\begin{table*}\center
\caption{Corpus Statistics \label{leng}}
\begin{tabular}{c|c|c|c|c|c|c|}
\cline{2-7}
& \multicolumn{2}{|c|}{No of Sentences} & \multicolumn{2}{|c|}{No of Words} & \multicolumn{2}{|c|}{Average sentence length}\\
\cline{2-7}
& Training & Testing & Training & Testing & Training & Testing \\
\hline
\multicolumn{1}{|c|} {English} & 4089 & 1090 & 110004 & 28381 & 26.90 & 26.03 \\
\hline
\multicolumn{1}{|c|} {Hindi} & 6096 & 1318 & 74897 & 30136 & 12.28 & 22.86\\
\hline
\multicolumn{1}{|c|} {Tamil} & 5422 & 1738 & 74594 & 25670 & 13.75 & 14.76 \\
\hline
\multicolumn{1}{|c|} {Malayalam} & 3287 & 1070 & 41716 & 13791 & 12.69 & 12.88 \\
\hline
\end{tabular}
\end{table*}
\begin{table}\center
\caption{Size of FIRE-2014 NER Data-set \label{Size_of_Dataset}}
\begin{tabular}{|c|c|c|c|}
\hline
\textbf{Languages} & \textbf{Training} & \textbf{Development} & \textbf{Test} \\
  & \textbf{Data} & \textbf{Data} & \textbf{Data} \\ \hline
\textbf{English} & 90005 & 19998 & 29473 \\ \hline
\textbf{Hindi} & 80992 & 13277 & 31453 \\ \hline
\textbf{Tamil} & 80015 & 17001 & 27407 \\ \hline
\textbf{Malayalam} & 45009 & 9782 & 14661 \\ \hline
\end{tabular}
\end{table}

\subsection{Data Description}
The training corpus of FIRE-2014 NER Track consists of 6 columns namely word, POS Tag, Chunk Tag and three levels of named entity tags. Additional features of linguistic features and binary features are also included in training corpus. Linguistic features differ for each language since they are language specific like Root words, context information, prefix and suffix information etc. Binary features are considered for all four language which include presence of dot, hyphen, parenthesis, presence of number etc.

\begin{table}\center
\caption{Number of unique Named Entity Tags in Corpus \label{UTag}}
\begin{tabular}{|c|c|c|c|}
\hline
Corpus & Level-1 & Level-2 & Level-3 \\
 & Tag & Tag & Tag \\
\hline
English &153 &75 &16 \\
\hline
Hindi &168 &61 & 8\\
\hline
Tamil & 166&62& 14\\
\hline
Malayalam & 158&69 & 14\\
\hline
\end{tabular}
\end{table}

The size of English corpus is large than other languages. Tamil and Hindi corpus size is almost nearer to English whereas Malayalam training corpus is only 50\% of the English corpus but all four languages have almost equal number of named entities. In all the languages 50\% of the corpus contains only Person and Location as label. The remaining corpus occupies the other tags.

In FIRE-2014, English NER system was alone developed using CRF and Indian Languages are developed using SVM algorithm. The training and development corpus of English language are clubbed together to form training set, so the size of training set increased from 90005 words to 110004 words. The Indian languages NER system uses development set for tuning the model.

The number of unique named entity tags present in training corpus is given in Table \ref{UTag}. The level-1 tags in training corpus for each language is more than 150 tags and level-2 tags are more than 60 in each language. Level-3 tags for English, Tamil and Malayalam are more than 10 whereas Hindi language alone has only 8 tags in level-3.

The test data is initially consisting of only three columns namely token or word, Parts-of-speech (POS) tag and Chunk tag. This testing corpus is also added with both linguistic features and binary features which are included while training. The un-tagged test data along with features are given for testing with three levels. The test data size of Malayalam language is also only 50\% of other languages.

\subsection{Performance Metrics}
Forum of Information Retrieval and Evaluation (FIRE-2014)- NER organizers evaluated the system for all the four languages based on Precision, Recall and Approximate match calculation \cite{42}. The evaluated results are given in the Table \ref{Result}.

Precision and recall are used for measuring the relevance of output retrieved from the system. Precision (P) and Recall (R) help in finding quality and quantity of the output respectively. Precision tells how many selected items are relevant whereas recall deals with how many relevant items are selected.
\begin{equation}
Precision(P)=relevant \; tags/retrieved \; tags
\end{equation}
\begin{equation}
Recall(R)=relevant \; tags/total \; possible \; relevant \; tags
\end{equation}
\begin{equation}
F_{1}= \frac{2.P\;R}{P+R}
\end{equation}

\textbf{Approximate Match Metric Calculation:}
\\
Approximate match metric is used for evaluating partial correctness of the named entity. The right boundary should match. The named entity tag should be same as the gold standard tag. The tags that are perfectly matched are given weightage of 1 and partially matched tags are given weightage of 0.5. Among 10 Named Entities identified by the system, if 4 are perfectly identified and 5 are partially identified then $approximate match=((4*1)+(5*0.5))/10=0.65$ \cite{53}.

\begin{table*}\center
\caption{Results evaluated by FIRE-2014 organizers of NER Track \label{Result}}
\begin{tabular}{|*6{c|}}
\hline
Language & & Precision & Recall & Approximate & F1\\
& & & & match & Measure \\
\hline
Hindi & Outer Level & 33.70 & 17.36 & 34.10 & 22.92 \\
\cline{2-6}
& Inner Level & 05.02 & 11.00 & 05.13 & 06.89\\
\hline
Malayalam & Outer Level & 27.931 & 21.89 & 27.93 & 24.54\\
\cline{2-6}
& Inner Level & 22.87 & 16.21 & 23.42 & 18.97 \\
\hline
Tamil & Outer Level	& 43.33 & 18.38 & 43.68 & 25.81\\
\cline{2-6}
& Inner Level & 38.04 & 12.28 & 38.043 & 18.56\\
\hline
English & Outer Level & 67.82 & 52.80 & 68.11 & 59.37\\
\cline{2-6}
& Inner Level & 66.42 & 32.26 & 66.42 & 43.42\\
\hline
Average	& Outer Level & 41.93 & 26.19 & 43.45 & 32.24\\
\cline{2-6}
& Inner Level & 33.25 & 19.36 & 34.25 & 24.47\\
\hline
\end{tabular}
\end{table*}

\begin{table*}\center
\caption{Improved Results after FIRE-2014 Evaluation \label{Result_1}}
\begin{tabular}{|*6{c|}}
\hline
Language & & Precision & Recall & Approximate & F1\\
& & & & match & Measure \\
\hline
Hindi & Outer Level & 38.98 & 26.18 & 39.98 & 31.32 \\
\cline{2-6}
& Inner Level & 08.30 & 25.00 & 08.47 & 12.46\\
\hline
Malayalam & Outer Level & 35.32 & 25.40 & 35.63 & 29.54\\
\cline{2-6}
& Inner Level & 39.05 & 17.83 & 39.05 & 24.48 \\
\hline
Tamil & Outer Level	& 54.32 & 28.19 & 55.04 & 37.11\\
\cline{2-6}
& Inner Level & 61.58 & 35.43 & 61.58 & 44.98\\
\hline
Average	& Outer Level & 42.87 & 26.59 & 43.55 & 32.82\\
\cline{2-6}
& Inner Level & 36.31 & 26.08 & 36.36 & 30.36\\
\hline
\end{tabular}
\end{table*}

\subsection{Results and Discussions}
The analysis of the nested NER system was done for outer and inner levels. The first level tags are outer level and second level tags are inner level tags. Based on the precision and recall evaluated by FIRE-2014, F1 measure was calculated. It shows that English NER system which was implemented using CRF shows better results than other Indian languages which were implemented using SVM. CRF consumes more time while training the model but increase the performance of the system. Among all languages, Hindi system shows very low performance for the inner level whereas outer level has equal performance with other two Indian languages. Malayalam and Tamil language shows almost same performance for their outer and inner level. Indian Languages show 50\% decreased performance than English. From this system, we observed there should be improvement in the model developed in order to increase the performance of all Indian languages.
The accuracy for all these mentioned languages are described in Table \ref{Result_1}. It shows that F1 measure of the Tamil language is increased from 25.81 to 37.11 in outer level and good performance increase for inner level from 18.56 to 44.98. Tamil language shows the high performance among the Indian languages which has approximate match metric of 55.04 for outer level and 61.58 for inner level.
The entities like numbers are identified correctly by these systems and tagged as COUNT, PERIOD, QUANTITY, DATE or TIME based on their occurrence. Systems are able to identify the numbers because of use of various binary features specific to number identification like presence of number, any digit number, two digit number, 3 digit number and 4 digit number. The most of the entities which represent place name or person name are also identified by the system because of use of gazetteer information. This system also fails to predict accurate tag. For example, while training the system the name of location, city and nation are tagged as LOC, CITY and NATION. This system fails to differentiate these three tags but any one of these are tagged for name of location, city or nation.

Table \ref{leng} gives the average length of sentence for all four languages. The average sentence length is almost same for training and testing in three languages namely English, Tamil and Malayalam whereas in Hindi language alone it differs in training and testing. Training corpus contains average length of 22.86 whereas testing has 12.88. This may also reduce the accuracy of Hindi language. In all comparisons, that the improved NER system for Indian languages using CRF gives better performance. The performance of all Indian languages are increased when compared to FIRE output. Based on the F1 measure, the average performance increase in all Indian languages for outer level is 33.59\% and the performance increase for inner level is about 83.07\%.
\subsection{NEEL - \#Micropost2015}
The experiment is conducted on i7 processor with 8GB RAM and the flow of experiment is shown in Figure \ref{Over}. The training dataset consists of 3498 tweets with the unique tweet id. These tweets have 4016 entities with 7 unique tags namely Character, Event, Location, Organization, Person, Product and Thing \cite{61}\cite{62}. POS tag for the NER is obtained from TwitIE tagger after tokenization which takes care of the nature of micro posts and provides an outcome desired by the POS tagger model. The tags are mapped to BIO Tagging of named entities. Considering the entity as a phrase, token at the beginning of the phrase is tagged as 'B-(original tag)' and the token inside the phrase is tagged as 'I-(original tag)'. Feature vector constructed with  POS tag and additional 34 features like root word, word shapes, prefix and suffix of length 1 to 4, length of the token, start and end of the sentence, binary features - whether the word contains uppercase, lower case, special symbols, punctuation, first letter capitalization, combination of alphabet with digits, punctuation and symbols, token of length 2 and 4 , etc.\\
After constructing the feature vector for individual tokens in the training set and by keeping bi-directional window of size 5, the nearby token's feature statistics are also observed to help the WSD. The final windowed training sets are passed to the CRF and SVM algorithms to produce the NER model. The development data has 500 tweets along with their id and 790 entities \cite{61}\cite{62}. The development data is also tokenized, tagged and feature extracted as the training data for testing and tuning the model.
\begin{figure}[!h]\centering
\includegraphics[scale=.5]{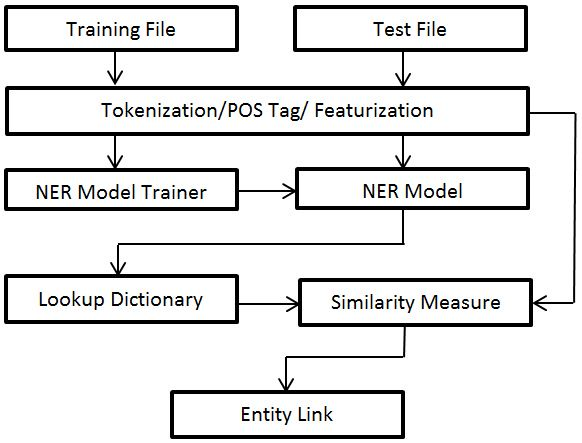}
\caption {Overall Model Structure \label{Over}}
\end{figure}
The developed model performance is evaluated by 10- fold cross validation of training set and validated against the development data. The accuracy is computed as ratio of total number of correctly identified entities to the total number of entities and tabulated in Table \ref{obs}. 
\begin{equation}
Accuracy = \frac{\sum correctly \; identified \; entities}{total \; entities} \times 100
\end{equation}
MALLET incorporates O-LBFGS which is well suited for log-linear models but shows reduced performance when compared to CRFsuite which engulfs LBFGS for optimization \cite{63}\cite{64}. SVM's low performance can be improved by increasing the number of features which will not introduce any over fitting and sparse matrix problem \cite{65}.\\
The final entity linking part is done by utilizing look-up dictionary (DBpedia 2014) and sentence similarity. The entity's tokens are given to the look up dictionary which results in few related links. The final link assigned to the entity is based on maximum similarity score between related links and proper nouns in the test tweet. Similarity score is computed by performing dot product between uni-gram vectors of proper nouns in the test tweet and the uni-gram vectors of related links from look-up dictionary. Entity without related links is assigned as NIL.
\begin{table}\center
\caption{Observations\label{obs}}
\begin{tabular}{|l|c|c|c|}
  \hline
\multicolumn{0}{|c|}{\textbf{Tools}} & \multicolumn{0}{|c|}{\textbf{10 Fold-Cross}}&\multicolumn{0}{|c|}{\textbf{Development}}&\multicolumn{0}{|c|}{\textbf{Time}} \\ 
&\textbf{Validation}&\textbf{Data}&\textbf{(mins)}\\\hline
Mallet & 84.9 & 82.4&168.31\\ \hline
SVM & 79.8 & 76.3&20.15\\ \hline
CRFSuite & 88.9 & 85.2&4.12 \\ \hline
\end{tabular}
\end{table}
This experimentation is about sequence labeling for entity identification from micro posts and extended with DBpedia resource linking. By observing Table 3.6, it is clear that CRF shows great performance and paves way for building a smart NER model for streaming data application. Even though CRF seems to be reliable, it is dependent on the feature that has direct relation with NER accuracy. The utilized TwitIE tagger shows promising performance in both the tokenization and POS tagging phases. The special 34 features extracted from the tweets improves efficacy by nearing 13\% greater than the model with absence of special features. At linking part, this work is limited using dot product similarity which could be improved by including semantic similarity.

\section{Co-Reference Resolution of both Singleton and Co-referent}
Co - Reference Resolution (CRR) stands and includes, solving challenging problems involved in  multiple applications in various domains, in which clinical domain applications considered here as an objective. The requirement of CRR induces the research and researcher to built the domain dependent and independent models, in order to solve the various problems. Though there were a pool of proposals based on unsupervised, supervised and rule based methods, it fails to fulfills the objective in multiple scenarios which cannot be entitled in clinical domain applications. By observing the problems, pros and cons in the existing methods, a novel approach Hierarchical Sequential Model (HSM) proposed here to solve the Co – Reference Resolution in clinical documents by solving both the Singleton and Co-Referent problems.\\
Co – Reference Resolution (CRR) known to be mapping pro-forms to their corresponding entities within a document. This involves multiple applications in clinical domain like question answering systems, Information Extraction, Topic Modelling, Sentence Simplification or splitting, Multi modal dialogue systems,  recommendation and recognition engines.\\
The objective introduces more complex problems when clinical document's nature joins with the existing Natural Language Processing (NLP) problems. Researches on general CRR started in mid of 19th century, though yet it remains as a research topic in various domains. Earlier,  unavailability of clinical data poses lag of research in  medical domain, this is conquered in recent years by utilizing Electronic Health Records (EHRs) and  Electronic Case Reports (ECRs). Hence this initiated and  induced much research works on clinical CRR based on heuristic approaches, supervised and unsupervised approaches.\\
Heuristic based approaches initiated to solve the problem and still many research work proposes and competes with the machine learning based methods. Generally in rule based methods, features like lexical expressions and term – syntactic feature pairs are assigned with weight, which is utilized to justify the defined rules but they involves conflicts while solving problem which as multiple answers.  To reduce the human effort and to increase the recall value for unseen categories corpus based approaches proposed where lexical, syntactic and semantic based features utilized to form the regression relations or classification rules.\\ 
Afterwards combination of above methods along with sentence parsing was carried out but yet the requirement stands as a problem.  Most of the above discussed systems has common disadvantages i.e. absence of relative information and tested on closed manual built dataset. Later Conditional Random Fields introduced to over come the above problem but it is not applied on clinical domain and they able to solve the singleton problem alone \cite{73}.\\
The key to solve the stated problem is to utilize the latent relational context information lied inside the document in which existing knowledge and data driven methods fails to incorporate \cite{74}. This can be achieved by solving the problem by utilizing sequential  models where the Conditional Random Fields (CRF), Markov Models (MM)  plays major role \cite{74}.  It is known that discriminative models like Maximum Entropy Markov Model (MEMM) and CRF outperforms  the generative model like Hidden Markov Model (HMM) due to the advantage of long range dependency. Further CRF shows promising outcome with reduced error rate (essential in clinical domain) when compared to the MEMM because of controlled label biasing problem. By accounting above mentioned requirement further approach carried out by utilizing CRF which involves long range dependency and avoids label biasing problem.
\section{Conclusion}
The NER system learned using CRF consumes more time while training the model. The corpus contains three levels of named entities in the case of FIRE, hence three levels separately used for training and testing. The dataset is not clean for Malayalam and Hindi language like we have Arabic words in between Malayalam words and English letters in between Hindi words in test data which may reduce the performance of the system. The POS tag is an important feature that helps in deciding the named entity. The dataset contains incorrect POS tag which reduced the accuracy. The proposed system which uses rich features like linguistic and binary features solves nested named entities using chain classifier. The specially extracted feature from tweets shows greater performance in NEEL task and CRF along with LBFGS optimization consumes less time while comparing with other. However the testing period is lows as compared to the NER along with SVM, which ensures CRF can be utilized in real time streaming applications. In this work the second level tagging is done by considering first level tag as one feature. This will be overcome by structured output learning in future works. Collocations and associative measures can be used as a feature to improve nested named entity recognition.

\chapter{Summary, conclusions and scope for further research}
Thus the NER and author profiling were done by having semantic information from the streaming data and batch data. In order to justify the experimented work and proposed approach among other related approaches, the data set are taken from the national and international shared tasks. The results of the NER are promising, which ensures to extend the further research in same. Comparing out come of the NER of English Language other languages shows reduced performance. This is because of the probability sparseness due to the morphological nature of the languages. This can be reduced by performing stemming, then it leads to be made compromise with the extraction of semantic information. Hence the feature work will be focused on solving probability sparseness without compromising the extraction of semantic information. The proposed Life Time Learning Model in author profiling shows greater performance than the systems that were ranked one at PAN 2013 and 2014 shared tasks. Hence the algorithm proves to be global model which is independent of language, topic and genre. It has to be noted that Life Time Learning Model even solves the Big Data problem, hence model can be projected has preprocessing stage of author profiling on Big Data in-order to make distributed computation. The proposed model proved to be there is no dependency among the batches, which ensures the scalability of the model. The only draw back about the model is time consumption, but it is negligible when the same model runs on the distributed computation framework. Hence the future work will be extending and implementing proposed algorithm on distributed computation frameworks like Apache Hadoop and Apache Spark. An assumption made that the space spanned by the feature vectors in individual batch has relation between space spanned by other batches. Future work will includes the proof to the assumption that is made.

\chapter*{List of publications based on the research work} \addcontentsline{toc}{chapter}{\hspace*{.53cm}List of publications based on the research work}
\begin{enumerate}
\item Barathi Ganesh HB, Abinaya N, Anand Kumar M, Vinayakumar R and Soman K P, "AMRITA - CEN@NEEL : Identification and Linking of Twitter Entities", World Wide Web Conference - 2015,  \#5th Workshop on Making Sense of Microposts, was held at Floarance, Italy on $18^{th}$ of May 2015. (Publisher: 'CEUR (ISSN 1613-0073)').
\item Barathi Ganesh HB, Reshma U and Anand Kumar M, "Author Identification based on Word Distribution in Word space", $4^{th}$ International Symposium on Natural Language Processing (NLP'15), will be held at SCMS, Aluva, Kochi between $10^{th}$ to $13^{th}$ of August 2015. (Publisher : 'IEEE Xplore').
\item Abinaya N, Neethu John, Barathi Ganesh HB, Anand Kumar M, Soman K P, "AMRITA@FIRE-2014: Named Entity Recognition for Indian Languages using Rich Features", working note in Forum for Information Retrieval and Evaluation (FIRE 2014), was held at $4^{th}$ of December 2014.
\end{enumerate}

\end{document}